\documentclass{article}

\usepackage{microtype}
\usepackage{graphicx}
\usepackage{subfigure}
\usepackage{amsmath}
\usepackage{amssymb}
\usepackage{booktabs} % for professional tables
\usepackage{tikz} % Required for flow chart
\usetikzlibrary{shapes,arrows.meta} % Tikz libraries required for the flow chart in the template
\usetikzlibrary{arrows,positioning}
\usepackage{hyperref}
\usepackage{xcolor}
\definecolor{darkgreen}{rgb}{0.035, 0.35, 0.145}

\tikzset{
	>=stealth',
	punkt/.style={
		rectangle,
		rounded corners,
		draw=black, very thick,
		text width=6.5em,
		minimum height=2em,
		text centered},
	pil/.style={
		->,
		thick,
		shorten <=2pt,
		shorten >=2pt,}
}

\DeclareMathOperator*{\argmin}{arg\,min}
\newcommand{\norm}[1]{\left\lVert#1\right\rVert}
\usepackage{multirow}
\usepackage{textcomp}

\renewcommand{\textdownarrow}{$\downarrow$}
\usepackage{rotating}
\usepackage{multicol}

\usepackage[accepted]{icml2020}
\graphicspath{{./Figures/}}
\icmltitlerunning{Black-box Adversarial Example Generation with Normalizing Flows}

\begin{document}

\twocolumn[
\icmltitle{Black-box Adversarial Example Generation with Normalizing Flows}




\begin{icmlauthorlist}
\icmlauthor{Hadi~M.~Dolatabadi}{to}
\icmlauthor{Sarah~Erfani}{to}
\icmlauthor{Christopher~Leckie}{to}
\end{icmlauthorlist}

\icmlaffiliation{to}{School of Computing and Information Systems, The University of Melbourne, Victoria, Australia}

\icmlcorrespondingauthor{Hadi~M.~Dolatabadi}{hadi.mohagheghdolatabadi@student.unimelb.edu.au}

\icmlkeywords{Normalizing Flows, Adversarial Example Generation, Black-box Adversarial Attack}

\vskip 0.3in
]



\printAffiliationsAndNotice{}  % leave blank if no need to mention equal contribution

\begin{abstract}
Deep neural network classifiers suffer from adversarial vulnerability: well-crafted, unnoticeable changes to the input data can affect the classifier decision.
In this regard, the study of powerful adversarial attacks can help shed light on sources of this malicious behavior.
In this paper, we propose a novel black-box adversarial attack using normalizing flows.
We show how an adversary can be found by searching over a pre-trained flow-based model base distribution.
This way, we can generate adversaries that resemble the original data closely as the perturbations are in the shape of the data.
We then demonstrate the competitive performance of the proposed approach against well-known black-box adversarial attack methods.
\end{abstract}

\section{Introduction}

Deep neural network (DNN) classifiers have been successfully applied to many object recognition tasks.
However, \citet{szegedy2014intriguing} pointed out that even the slightest intentional changes to a DNN input, widely known as \textit{adversarial attacks}, can change the classifier decision.
This observation is peculiar as those changes are tiny and can barely affect a human's judgment about the object class.
Since their emergence, many adversarial attack methods have been devised.
These studies are often helpful in recognizing sources of this misbehavior, which ultimately can lead to more robust DNN classifiers.

There are many attributes by which adversarial attacks can be categorized~\citep{yuan2019adversarial}.
Perhaps the most famous one is with respect to the adversary's knowledge about the target DNN.
In this sense, threat models are divided into white- and black-box attacks.
In \textit{white-box} attacks, it is assumed that the adversary has full access to the internal weights of the target DNN, and can leverage this knowledge in generating adversarial examples by using the DNN gradients.
In contrast, \textit{black-box} adversarial attacks are assumed to have access to solely the input and output of a classifier.
As a result, they have to utilize this limited capacity in order to construct their adversarial examples.

There has been some research on the use of generative models in the construction of adversarial examples, for instance, ~\citet{baluja2018learning, xiao2018generating, song2018constructing, wang2019adirect}.  
These works are mostly concerned with training a generative model on a target network so that the samples generated by them are adversarial.
To this end, they often require taking the gradient of the target network, and hence, are mostly suitable for white-box settings.
To adapt themselves to the black-box scenario, they often replace the target network with a substitute version.
Thus, the performance of such approaches heavily depends on the resemblance of the target network to the substitute one.
Moreover, for various types of defenses, such methods often require re-training their generator on a different substitute network.

In this paper, we propose using pre-trained flow-based models to generate adversarial attacks for the black-box setting.
We first formulate the problem of adversarial example generation.
Then, we show how searching over the base distribution of a pre-trained normalizing flow can be related to generating adversaries.
Finally, we show the effectiveness of the proposed method in attacking vanilla and defended models.
We see that the perturbations generated by our method follow the shape of the data closely. However, this is generally not the case for other existing methods, as they often look like an additive noise.

To the best of our knowledge, this is the first work that exploits normalizing flows to generate adversarial examples.
Through the experimental results, we see how this method can be used to make adversarial perturbations less noticeable.
We hope our work can be a stepping stone into modeling adversaries using exact likelihood approaches with their ability to model the data distribution closely.
Hopefully, such works can lead to the statistical treatment of DNNs' adversarial vulnerability.

\section{Background}

\subsection{Normalizing Flows}

Normalizing flows~\citep{tabak2013family, dinh2015nice, rezende2015variational} are a relatively novel family of generative models.
They use invertible neural networks (INN) to transform a simple density into data distribution.
To this end, they exploit the \textit{change of variables} theorem.
In particular, assume that ${\mathbf{Z} \in \mathbb{R}^{d}}$ denote an arbitrary random vector from a uniform or standard normal distribution.
If we construct a new random vector ${\mathbf{X} \in \mathbb{R}^{d}}$ by applying a differentiable INN ${\mathbf{f}(\cdot): \mathbb{R}^{d} \rightarrow \mathbb{R}^{d}}$ to ${\mathbf{Z}}$, then the relationship between their corresponding densities can be written as
\begin{equation}\label{eq:change_of_variable}
p(\mathbf{x}) = p(\mathbf{z})\left|\mathrm{det}\Big(\dfrac{\partial \mathbf{f}}{\partial \mathbf{z}}\Big)\right|^{-1}.
\end{equation}
The multiplicative term on the RHS is known as the \textit{Jacobian determinant}.
This term accounts for \textit{normalizing} the base distribution ${p(\mathbf{z})}$ such that the density ${p(\mathbf{x})}$ represents the data distribution.
To make modeling of high-dimensional data feasible, the Jacobian determinant must be computed efficiently.
Otherwise, this calculation can hinder the application of such models to high-dimensional data as the cost of computing the determinant grows cubically with the data dimension.
Once set, we can use \textit{maximum likelihood} to fit the flow-based model of Eq.~\eqref{eq:change_of_variable} to data observations.
This fitting is done using numerical optimization methods such as Adam~\citep{kingma2015adam}.

One of the earliest INN designs for flow-based modeling is Real~NVP~\citep{dinh2016density}.
This network uses affine transformations in conjunction with ordinary neural networks such as ResNets~\citep{he2016deep} to construct a normalizing flow.
In this paper, we use a reformulation of Real~NVP~\citep{dinh2016density} introduced by \citet{ardizzone2019guided}.
This transformation is defined by stacking two consecutive layers of ordinary Real~NVP together
\begin{align}\label{eq:realnvp_ardizzone}
\mathbf{x}_1 &= \mathbf{z}_1 \odot \exp\big(\mathbf{s}_1({\mathbf{z}_2})\big) + \mathbf{t}_1({\mathbf{z}_2}) \nonumber \\
\mathbf{x}_2 &= \mathbf{z}_2 \odot \exp\big(\mathbf{s}_2({\mathbf{x}_1})\big) + \mathbf{t}_2({\mathbf{x}_1}).
\end{align}
Here, ${\mathbf{s}_{1,2}(\cdot)}$ and ${\mathbf{t}_{1,2}(\cdot)}$ represent the scaling and translation functions, and they are implemented using ordinary neural networks as they are not required to be invertible.
For more information about flow-based models and architectures, we refer the interested reader to~\citet{kobyzev2019nfs, papamakarios2019normalizing}.   

\subsection{Adversarial Example Generation}
 
Let $\mathcal{C}(\cdot)$ denote a DNN classifier.
Assume that this network is defined so that it takes an image $\mathbf{x}$ as its input, and outputs a vector whose $y$-th element indicates the probability of the input belonging to class $y$.
Now, we can solve the following optimization problem to find an adversarial example for $\mathbf{x}$
\begin{align}\label{eq:CW_loss}
{\mathbf{x}}_{adv}=\argmin_{\norm{\mathbf{x}' - \mathbf{x}}_{p}\leq \epsilon} \mathcal{L}(\mathbf{x}').
\end{align}
Here, ${\mathcal{L}(\mathbf{x}')=\max\big(0, \log \mathcal{C}(\mathbf{x}')_{y} - \max_{c \neq y} \log \mathcal{C}(\mathbf{x}')_{c}\big)}$ is the \citet{carlini2017towards} (C\&W) loss.
This objective function is always non-negative.
Upon becoming zero, it indicates that we have found a category for which the classifier outputs a higher probability than the data, and hence, constructed an adversarial example.
Moreover, we limit our search to the images whose $\ell_p$ norm lies within the \mbox{$\epsilon$-boundary} of the original image.
This constraint is in place to ensure that the adversarial image looks like the clean data.

White-box attacks can leverage the network architecture and internal weights to solve the objective of Eq.~\eqref{eq:CW_loss} via back-propagating through the classifier ${\mathcal{C}(\cdot)}$.
However, in black-box attacks, we are restricted to querying the classifier ${\mathcal{C}(\cdot)}$ and working with its outputs only.
In this paper, we are going to solve Eq.~\eqref{eq:CW_loss} for an adversarial image in the black-box setting. 

\section{Proposed Method}

Consider a flow-based model that is trained on some image dataset in an unsupervised manner.
It was empirically shown that given such a generator, all the latent points in a neighborhood tend to generate visually similar pictures.
This property is the result of the invertibility and differentiability of normalizing flows, which causes the image manifolds to be smooth~\citep{kingma2018glow}.
We can exploit this property of flow-based models to generate adversarial examples.
To this end, we need to search in the vicinity of the latent representation of an image, and find the one that minimizes the cost function of Eq.~\eqref{eq:CW_loss}.
We can achieve this goal by assuming an adjustable base distribution around a given image's latent representation.
Then we tune this base distribution so that it generates an adversarial example.
The natural way of doing so is to consider an isometric Gaussian with non-zero mean as the base distribution of the normalizing flow, as opposed to the standard Gaussian, which is used in training it.

\begin{figure*}[t!]
	\centering
	\begin{tikzpicture}[node distance = 1cm]
	\node [inner sep=0pt] at (0,0) (orig) {\includegraphics[width=.1\textwidth]{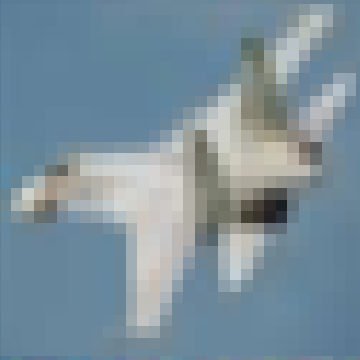}};
	\node[text width=3cm] at (0.4,-1.25) {Clean Image $\mathbf{x}$};

	\draw [dashed, line width=1pt] (5,0) circle (1.5cm);
	\node[draw, color=darkgreen!100, fill=darkgreen!35, very thick, circle, radius=.2, name=latorig] at (4.2,0.3) {};
	
	\filldraw[black] (5,0) circle (1.25pt) node[anchor=west] {};
	\node[draw, color=red!100, fill=red!35, very thick, circle, radius=.2, name=latadv1] at (4,1) {};
	\node[draw, color=red!100, fill=red!35, very thick, circle, radius=.2, name=latadv2] at (5.,1.4) {};
	\node[draw, color=red!100, fill=red!35, very thick, circle, radius=.2, name=latadv3] at (4.41,-1.2) {};
	\node[draw, color=red!100, fill=red!35, very thick, circle, radius=.2, name=latadv4] at  (5.7,-1.) {};
	\node[draw, color=red!100, fill=red!35, very thick, circle, radius=.2, name=latadv5] at (6.4,0.5) {};
	\node[draw, color=red!100, fill=red!35, very thick, circle, radius=.2, name=latadv6] at (3.6,-0.5) {};
	\node[draw, color=red!100, fill=red!35, very thick, circle, radius=.2, name=latadv7] at (6.5,-0.5) {};
	
	\node [inner sep=0pt] at (10,1.5) (adv2) {\includegraphics[width=.1\textwidth]{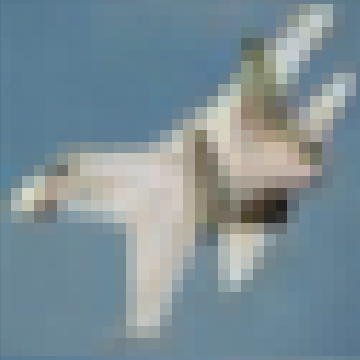}};
	\node [inner sep=0pt] at (10,0.5) (adv3) {\includegraphics[width=.1\textwidth]{adv.png}};
	\node [text width=3cm, rotate=90] at (10,0.5) {$\mathbf{\cdots}$};
	\node [inner sep=0pt] at (10,-2) (adv6) {\includegraphics[width=.1\textwidth]{adv.png}};
	\node[text width=3cm] at (10.4,-3.25) {Adv. Images $\mathbf{x}_{adv}$};
	
	\path[->, thick] (orig.east) edge[bend left=17.] node[above] {$\mathbf{f}^{-1}(\cdot)$} (latorig.west);
	\path[->, thick] (4.2,0.3) edge node[rotate=-35,below] {$\boldsymbol{\mu}$} (4.95,0.03);
	\path[->, thick] (5,0.0) edge node[rotate=45,above] {$\sigma\sqrt{d}$} (6.07, 1.07);
	
	\node[text width=3cm] at (8,-1.5) {$\mathbb{R}^{d}$};
	
	\path[->, thick] (7, 0) edge node[above] {$\mathbf{f}(\cdot)$} (8.75, 0);
	\path[->, thick] (11.25, 0) edge node[above] {\scriptsize${\norm{\mathbf{x}_{adv} - \mathbf{x}}_{p}\leq \epsilon}$} (13.5, 0);
	\node[text width=5cm, rotate=90] at (14.25, 0) {\small Select top-$k$ samples with the lowest C\&W loss, and set ${\boldsymbol{\mu}}$ to the average of their base representation ${\mathbf{z}}_{adv}$.};

	\end{tikzpicture}
	\caption{Adversarial example generation with pre-trained flow-based model $\mathbf{f}(\cdot)$.}
	\label{fig:adv}
\end{figure*}

In particular, let ${\mathbf{f}(\cdot)}$ denote our pre-trained normalizing flow.
Furthermore, let ${\mathbf{z}_{clean}=\mathbf{f}^{-1}(\mathbf{x}_{clean})}$ be the base distribution representation of the clean test image ${\mathbf{x}_{clean}}$.
Given the smoothness property of the generated images manifold, we assume that the adversarial example is being generated from
\begin{align}\label{eq:z_adv}
{\mathbf{z}}_{adv}={\mathbf{z}}_{clean} + \boldsymbol{\mu} + \sigma \boldsymbol{\epsilon}
\end{align}  
on the latent space of the flow-based model.
Here, $\boldsymbol{\mu} \in \mathbb{R}^{d}$ and $\sigma \in \mathbb{R}$ are the parameters that control the movement of our algorithm in the base distribution space.
We set $\sigma \in \mathbb{R}$ via hyper-parameter tuning, and keep $\boldsymbol{\mu}$ as an adjustable parameter in our algorithm. 
Furthermore, we assume ${\boldsymbol{\epsilon}}$ to come from a standard normal distribution.
In other words, Eq~\eqref{eq:z_adv} defines a vicinity of the target image $\mathbf{x}$ in the base distribution space.
We then try to adjust the positioning of this distribution through parameter $\boldsymbol{\mu}$  so that it generates adversarial examples.

In order to generate an adversarial example, we propose the following iterative algorithm.
First, we initialize $\boldsymbol{\mu}$ to a small random vector.
Next, $n$ samples of ${\mathbf{z}}_{adv}$ are drawn according to Eq.~\eqref{eq:z_adv}.
These samples are then translated into their corresponding images using the pre-trained flow-based model $\mathbf{f}(\cdot)$.
Afterward, we compute the C\&W loss for all of these samples by querying the target DNN $\mathcal{C}(\cdot)$.
Out of these samples, we select the top-$k$ ones for which the C\&W objective is the lowest.
We then update the vector $\boldsymbol{\mu}$ by averaging over the base distribution representation of the $k$~chosen samples that result in the lowest C\&W costs.
This procedure is repeated until we reach an adversarial example or hit the quota for the maximum number of classifier queries. 
Note that in order to satisfy ${\norm{\mathbf{x}_{adv} - \mathbf{x}}_{p}\leq \epsilon}$, we have to project the generated data into their corresponding images for which they satisfy this constraint in each iteration.
Figure~\ref{fig:adv} shows a schematic of the proposed framework.

A key advantage of our proposed method is that the adversarial perturbations found lie on the image manifold, and hence, should reflect the structure of the clean image.
This property is in contrast to traditional methods whose perturbations do not necessarily follow the image manifold. 

\section{Experiments}

To evaluate our proposed method, we first train a flow-based model on the training part of the CIFAR-10~\citep{krizhevsky2009learning} dataset.
To this end, we use the framework of \citet{ardizzone2019guided} for invertible generative modeling.\footnote{\href{https://github.com/VLL-HD/FrEIA}{github.com/VLL-HD/FrEIA}}
We use a two-level architecture for our normalizing flow.
At level one, the data first goes through $4$ layers of modified Real-NVP~(Eq.~\eqref{eq:realnvp_ardizzone}).
We then reduce the image resolution using RevNet downsamplers~\citep{jacobsen2018irevnet}.
Next, the image is sent through $6$ layers of low-resolution invertible mappings.
In this first level, all the transformations exploit convolutional neural networks.
Afterward, three-quarters of the data is sent directly to the output.
The rest then goes under another round of transformations that consists of $6$ fully-connected layers.
Table~\ref{tab:nf_details} in the Appendix summarizes the hyperparameters used for training the flow-based part of our black-box attack.

Note that although here we use Real~NVP~\citep{dinh2016density} as our flow-based model, we are not restricted to use this method.
In fact, any other normalizing flow that has an easy-to-compute inverse (such as NICE~\citep{dinh2015nice}, Glow~\citep{kingma2018glow}, and spline-based flows~\citep{mueller2019neural, durkan2019cubic, durkan2019neural, dolatabadi2020lrs})
can be used within our approach.

\begin{figure*}[tb!]
	\centering
	\includegraphics[width=0.8\textwidth]{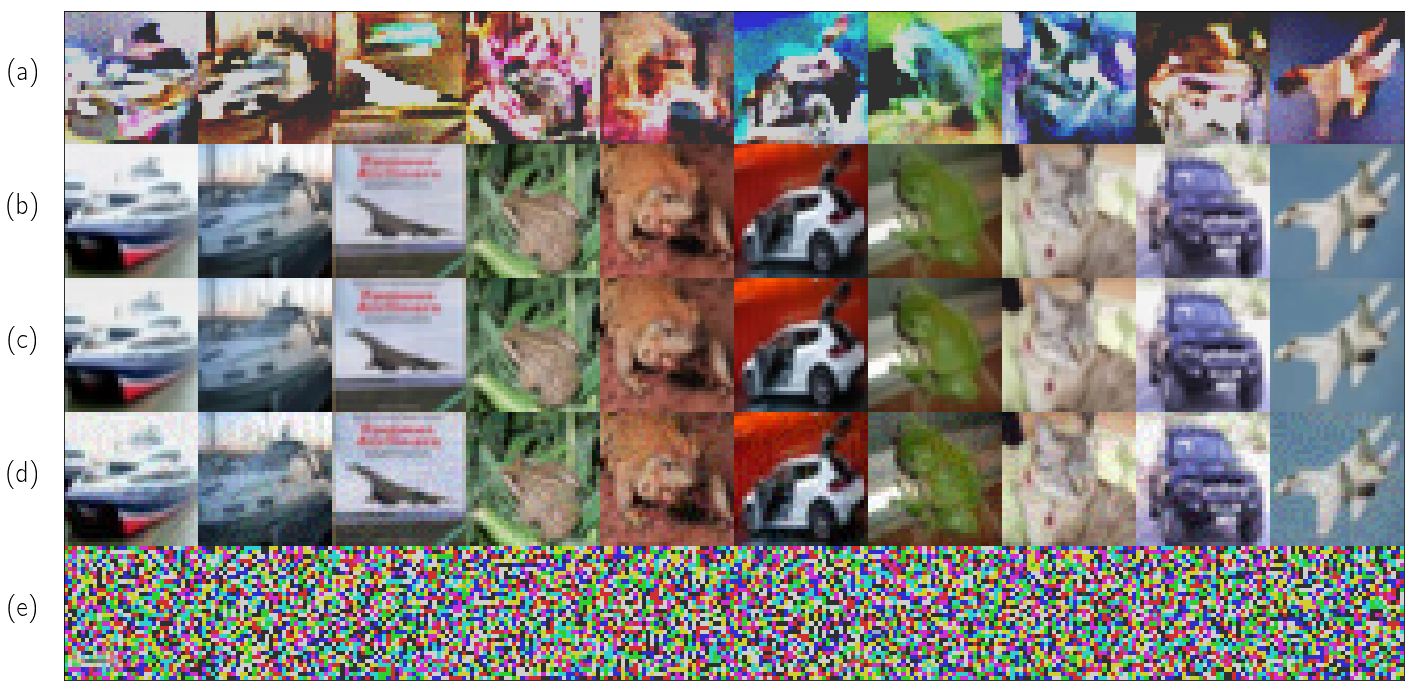}
	\caption{Magnified perturbation and adversarial examples generated by our proposed method vs. bandit attacks~\citep{ilyas2019prior}. As can be seen, the proposed method generates more realistic adversaries by using a flow-based model as its prior. (a) our method magnified perturbation (b) our method adversarial example (c) clean image (d) bandits adversarial example (e) bandits magnified perturbation.}
	\label{fig:adv_imgs}
\end{figure*}

\begin{table*}[tb!]
	\caption{Attack success rate (in \%, higher is better) to generate an adversarial example for CIFAR-10~\citep{krizhevsky2009learning} test data.
		Clean data accuracy and white-box PGD-100 attack success rate are also shown as a reference.
		All attacks are with respect to the $\ell_{\infty}$ norm with $\epsilon=8/255$.}
	\label{tab:success}
	\vskip 0.05in
	\begin{center}
			\resizebox{!}{!}{
				\begin{tabular}{cccccc}
					\toprule
					& &\multicolumn{4}{c}{Attack}\\
					\cmidrule(lr){3-6}
					Defense & Clean Acc.(\%)     & {PGD-100}  & \multicolumn{1}{c}{NES}   & \multicolumn{1}{c}{Bandits} & \multicolumn{1}{c}{Flow-based (ours)}\\
					\midrule
					Vanilla                      & $91.77$    & $100$   & $99.53$ & $98.68$ & $99.12$\\
					FreeAdv                      & $81.29$    & $47.52$ & $23.45$ & $37.10$ & $41.06$\\
					FastAdv                      & $86.33$    & $46.37$ & $20.15$ & $36.70$ & $40.06$\\
					RotNetAdv                    & $86.58$    & $46.59$ & $20.64$ & $36.67$ & $40.50$\\
					\bottomrule
			\end{tabular}}
	\end{center}
	\vskip -0.1in
\end{table*}

\begin{table*}[tb!]
	\caption{Average and median of the number of queries used to generate an adversarial example for scenarios of Table~\ref{tab:success}.}
	\label{tab:query}
	\vskip 0.05in
	\begin{center}
			\resizebox{!}{!}{
				\begin{tabular}{ccccccc}
					\toprule
					& \multicolumn{3}{c}{Avg. of Queries \textdownarrow} & \multicolumn{3}{c}{Med. of Queries \textdownarrow}\\
					\cmidrule(lr){2-4} \cmidrule(lr){5-7}
					Defense                      & NES & Bandits & Flow-based (ours)   & NES & Bandits & Flow-based (ours)\\
					\midrule
					Vanilla                      & $458.50$  & $524.14$  & $991.98$   & $300$ & $156$ & $460$\\
					FreeAdv                      & $629.38$  & $1430.30$ & $842.37$   & $100$ & $463$ & $180$\\
					FastAdv                      & $1465.51$ & $1425.78$ & $904.78$   & $800$ & $454$ & $200$\\
					RotNetAdv                    & $1526.46$ & $1470.35$ & $821.80$   & $800$ & $520$ & $180$\\
					\bottomrule
			\end{tabular}}
	\end{center}
	\vskip -0.1in
\end{table*}

Next, we select a Wide-ResNet-32~\citep{zagoruyko2016wresnet} with width $10$ as our classifier architecture.
This classifier is trained in both vanilla and defended fashions.
For the defended case, we use \textit{free}~\citep{shafahi2019free} and \textit{fast}~\citep{wong2020fast} adversarial training alongside \textit{adversarial training with auxiliary rotations}~\citep{hendrycks2019using}.
Each one of these classifiers is trained with respect to the $\ell_{\infty}$ norm with $\epsilon=8/255$.

Once the training is done, we can then perform our proposed black-box adversarial attack.
To this end, we try to generate an adversary from CIFAR-10 unseen test data.
An attack is counted successful if it can change the classifier decision about a correctly classified image in less than $10,000$ queries.
We compare our method against NES~\citep{ilyas2018black} and bandits with time and data-dependent priors~\cite{ilyas2019prior}.
The hyperparameters of each method are given in Tables~\ref{tab:ap:NES_details}-\ref{tab:ap:AdvFlow_details} in the Appendix.

Tables~\ref{tab:success} and~\ref{tab:query} show the attack success rate as well as the average and median of the number of queries for attacking nominated DNN classifiers.
As can be seen, the proposed method can improve the performance of baselines in attacking defended classifiers in both attack strength (success rate) and efficiency (number of queries).
Also, we see that the number of required queries for the proposed method remains almost consistent for both vanilla and defended classifiers.
However, this is not generally the case for the other methods, and their performance heavily depends on the classifier type.
Furthermore, as shown in Figure~\ref{fig:adv_imgs}, the adversarial examples generated by the proposed method look less suspicious in contrast to bandit attacks~\citep{ilyas2019prior}.
Also, we see that the perturbations generated by our approach are disguised in the underlying image structure.
However, bandit attack~\citep{ilyas2019prior} perturbations do not have this property and look like an additive noise.

\section{Conclusion}

In this paper, we proposed a novel black-box adversarial attack method using normalizing flows.
In particular, we utilize a pre-trained flow-based model to search in the vicinity of the base distribution representation of the target image and generate an adversarial example.
Due to the smoothness of image manifolds in normalizing flows, our adversarial examples look natural and unnoticeable.
This way, we can generate adversaries that can compete with well-known methods in terms of strength and efficiency.
We hope that this work can be inspiring in exploiting such methods for adversarial machine learning and lead to finding statistical treatments to DNNs' adversarial vulnerabilities.

\section*{Acknowledgements}
 This research was undertaken using the LIEF HPC-GPGPU Facility hosted at the University of Melbourne. This Facility was established with the assistance of LIEF Grant LE170100200.


\bibliography{references}
\bibliographystyle{icml2020}

\vfill\eject
\appendix
\textbf{\Large Appendix}
\section{Experimental Settings}
\begin{table}[htp]
	\caption{Hyperparameters used in training flow-based part of our approach.}
	\label{tab:nf_details}
	\vskip 0.1in
	\begin{center}
		\begin{small}
			\begin{tabular}{lc}
				\toprule
				Optimizer                          & Adam\\
				Scheduler                          & Exponential\\
				Initial learning rate              & $10^{-4}$\\
				Final learning rate                & $10^{-6}$\\
				Batch size                         & $64$\\
				Epochs                             & $350$\\
				\midrule
				Multi-scale levels                 & $2$\\
				Each level network type            & CNN-FC\\
				High-res transformation blocks     & $4$\\
				Low-res transformation blocks      & $6$\\
				FC transformation blocks           & $6$\\
				$\alpha$ (clamping hyperparameter)& $1.5$\\
				CNN layers hidden channels         & $128$\\
				FC layers internal width           & $128$\\
				Activation function                & Leaky ReLU\\
				Leaky slope                        & $0.1$\\
				\bottomrule
			\end{tabular}
		\end{small}
	\end{center}
\end{table}

\begin{table}[htp]
	\caption{Hyperparameters of query-limited NES attack~\citep{ilyas2018black}.}
	\label{tab:ap:NES_details}
	\vskip 0.15in
	\begin{center}
		\begin{small}
			\begin{tabular}{lcc}
				\toprule
				Hyperparameter          & Vanilla   & Defended\\
				\midrule
				$\sigma$ (noise std.)   & $0.1$     & $0.001$\\
				Sample size             & $50$      & $100$ \\
				Learning rate           & $0.01$    & $0.01$\\
				\bottomrule
			\end{tabular}
		\end{small}
	\end{center}
\end{table}

\begin{table}[htp]
	\caption{Hyperparameters of bandits with time and data-dependent priors~\cite{ilyas2019prior}.}
	\vskip 0.15in
	\label{tab:ap:bandits_details}
	\begin{center}
		\begin{small}
			\begin{tabular}{lcc}
				\toprule
				Hyperparameter                     & Vanilla              & Defended\\
				\midrule
				OCO learning rate                  & $100$                & $0.1$\\
				Image learning rate                & $0.01$               & $0.01$\\
				Bandit exploration                 & $0.1$                & $0.1$\\
				Finite difference probe            & $0.1$                & $0.1$\\
				Tile size                          & $(6\mathrm{px})^{2}$ & $(4\mathrm{px})^{2}$\\
				\bottomrule
			\end{tabular}
		\end{small}
	\end{center}
\end{table}

\begin{table}[ht!]
	\caption{Hyperparameters of our flow-based adversarial attack.}
	\label{tab:ap:AdvFlow_details}
	\vskip 0.15in
	\begin{center}
		\begin{small}
			\begin{tabular}{lc}
				\toprule
				Hyperparameter                                   & Value\\
				\midrule
				$\sigma$ (noise std.)                            & $0.1$\\
				Sample size                                      & $20$\\
				$k$ (samples used to update $\boldsymbol{\mu}$)  & $4$\\
				Maximum iteration                                & $500$\\
				\bottomrule
			\end{tabular}
		\end{small}
	\end{center}
\end{table}
\vfill
\null
%%%%%%%%%%%%%%%%%%%%%%%%%%%%%%%%%%%%%%%%%%%%%%%%%%%%%%%%%%%%%%%%%%%%%%%%%%%%%%%
%%%%%%%%%%%%%%%%%%%%%%%%%%%%%%%%%%%%%%%%%%%%%%%%%%%%%%%%%%%%%%%%%%%%%%%%%%%%%%%

\end{document}